\documentclass[]{sig-alternate}
\usepackage{cite}      

\usepackage{graphicx}  
\newcommand{\commentaire}[1]{}
\def\sharedaffiliation{
\end{tabular}
\begin{tabular}{c}}

\def\more-auths{
\end{tabular}
\begin{tabular}{c}}
\begin{document}
\title{Anisotropic Selection in Cellular Genetic Algorithms}
\numberofauthors{3}            
\author{
%
\alignauthor
\alignauthor David Simoncini\\
 \email{David.Simoncini@unice.fr}
\alignauthor
\more-auths
\alignauthor S\'ebastien Verel\\
 \email{Sebastien.Verel@unice.fr}
\alignauthor Philippe Collard\\
 \email{Philippe.Collard@unice.fr} 
\alignauthor Manuel Clergue\\
 \email{Manuel.Clergue@unice.fr}
       \sharedaffiliation
	   \affaddr{I3S Laboratory, CNRS-Universit\'e de Nice Sophia Antipolis}
}
\conferenceinfo{GECCO'06,} {July 8--12, 2006, Seattle, Washington, USA.}
\CopyrightYear{2006}
\crdata{1-59593-186-4/06/0007}

\maketitle

\begin{abstract} 
In this paper we introduce a new selection scheme in cellular genetic
algorithms (cGAs).
\textit{Anisotropic Selection} (AS) promotes diversity and allows
accurate control of the selec\-tive pressure. 
First we compare this new scheme with the classical
rectangular grid shapes solution according to the selective pressure: 
we can obtain the same takeover time with the two techniques although the spreading of the best individual is different.
We then give experimental results that
show to what extent AS promotes the emergence of niches that support low coupling and high cohesion. Finally, using a cGA with anisotropic selection on a Quadratic Assignment Problem we show the existence of an anisotropic optimal value for which the best average performance is observed.
Further work will focus on the selective pressure self-adjustment ability provided by this new selection scheme.            
\end{abstract}
\category{I.2.8}{Problem Solving, Control Methods, and Search}[Heuristic methods]
\terms{Algorithms}
\section*{Introduction}
This paper deals with selective pressure and diversity
 in cellular genetic algorithms (cGAs) which are a subclass of
 Genetic Algorithms where the population is embedded in a grid.
These concepts are closely related to the explo\-ration/exploitation trade-off. 
Previous studies on cGAs se\-lected the size and the shape of neighborhoods \cite{SarmaJ96},
 or the shape of the grid \cite{AlbaT00,Giacobini2003,Giacobini2004} 
as basic parameters to tune the search process. 
Altering these structural parameters entails a deep change in the way we deal with the problem.
 For instance, there is no built-in mean to swap from a rectangular grid 
to a square grid without misshaping the neighborhood relation.
 We suggest using \textit{anisotropic selection} (AS) to promote 
diversity and to control accurately the selective pressure in genetic search. 
The main advantage of the \textit{anisotropic selection} scheme is that it allows to
 control the exploration/exploitation trade-off without affecting neither the grid
 topology nor the neighborhood shape; so the cellular genetic algorithm we
 propose merely works on a square grid and a simple Von Neumann neighborhood shape.\\
 The paper is divided in 6 sections. Section \ref{section1} gives a brief definition of cGAs and
 an overview of existing techniques used to control the exploration/exploitation tradeoff. 
Section \ref{section2} introduces the AS scheme. Section \ref{section3} studies the
 influence of AS on the selective pressure. In Section \ref{section4} we compare AS 
and rectangular grids topologies' influence on the selective pressure. 
In Section \ref{section5} we show how AS promotes the emergence of niches. In Section
 \ref{section6} we use a cGA 
on a Quadratic Assignment Problem to measure the correlation between anisotropy and performance.
Finally we tie together the results of the previous sections and suggest directions
 for further research.   

\section{Selection in Cellular Genetic Algorithms}

This section presents a brief overview on cellular Genetic Algorithms and
 a standard technique to measure the 
selective pressure.

\label{section1}
\subsection{Cellular Genetic Algorithms}

Cellular Genetic Algorithms are a subclass of Genetic Algorithms (GAs) in
 which exploration and population diversity are enhanced thanks to the existence of
 small overlapped neighborhoods \cite{SpiessensM91}. Such algorithms are
 specially well suited for complex problems \cite{JongS95}.
We assume a two-dimensional toroidal grid as a spatial population structure. Each
 grid cell contains one individual of the population. The overlapping neighborhoods 
provide an implicit mechanism for migration of genetic material throughout the grid. 
A genetic algorithm is assumed to be running simultaneously on each grid cell, continuously
 selecting parents from the neighborhood of that grid cell in order to produce an
 offspring which replaces the current individual.

\subsection{Takeover Time}
\label{sec-takRG}

A standard technique to study the induced selection pressure without introducing the
perturbing effect of variation operators
is to let selection be the only active operator, and then monitor the number of best
 individual copies $N(t)$ in the population \cite{GoldbergD90}. The takeover time is the time
it takes for the single best individual to conquer the whole population. 
The grid is initialized with one cell having the best fitness and all the other
 having a null fitness. 
Since no other evolution mechanism but selection takes place, we can observe the way the best 
individual spreads over the grid by counting generation after generation the number 
of copies of this one.
A shorter takeover time thus means a higher selective pressure. 
It has been shown that
when we move from a panmictic population, as in standard GA, to a spatially structured one of the
same size with synchronous updating of the cells, the global selection pressure
induced on the entire population is weaker \cite{SarmaJ96}.

Links have been established between takeover time and neighborhood size 
and shape or grid topology.
Neighborhood size and shape in a cGA are parameters that have some influence on the
 takeover time. 
A larger overlap of local neighborhoods of the same shape speeds up the best 
individual's spreading over the grid. 
The influence of the shape is given by Sarma and De Jong through a measure on
 the neighborhood which represents the spatial dispersion of a cell pattern \cite{SarmaJ96}. 
Rather than the size of the neighborhood in terms of individuals, the key particularity
 of a local neighborhood is its radius. The takeover time decreases while the radius
 increases in a spatially structured population.

\begin{figure}[ht!]

\begin{center}
\begin{tabular}{c}
\includegraphics[width=6cm,height=6cm]{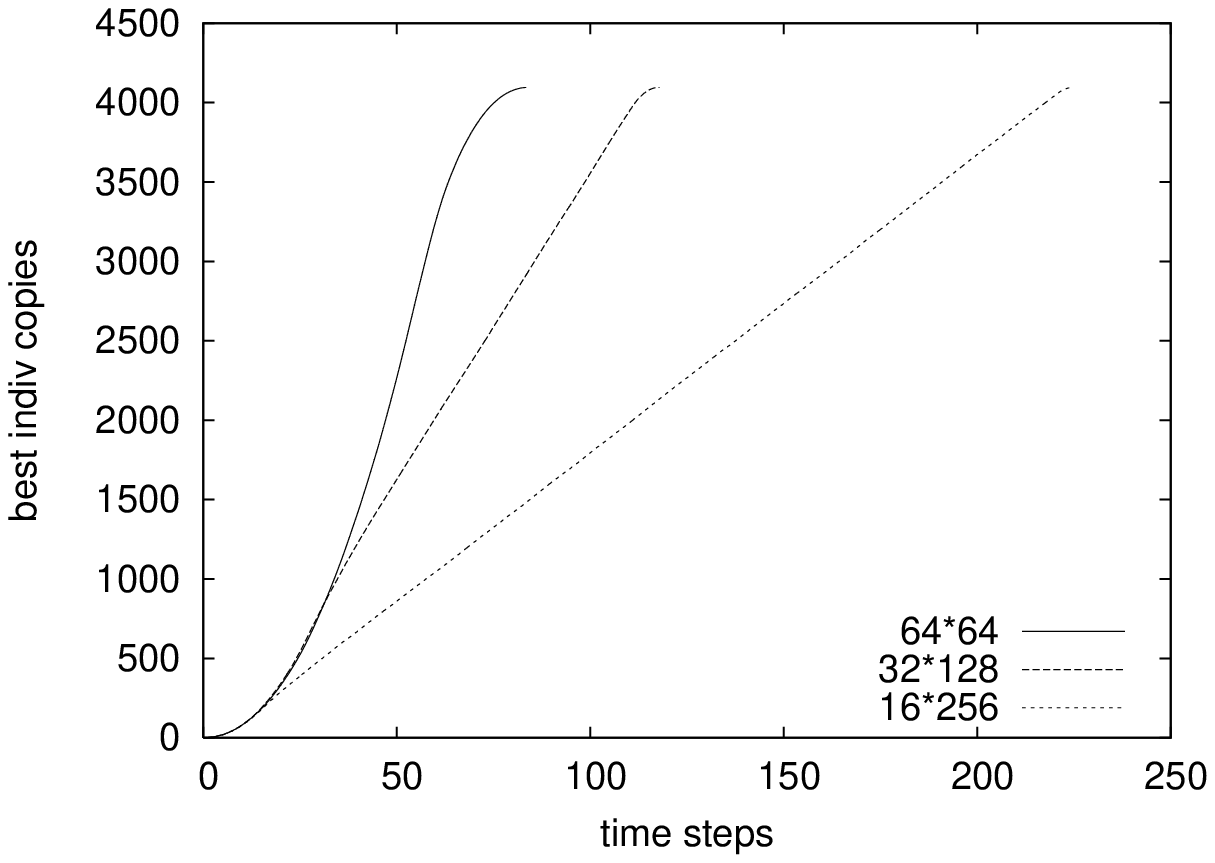} \\
\includegraphics[width=6cm,height=6cm]{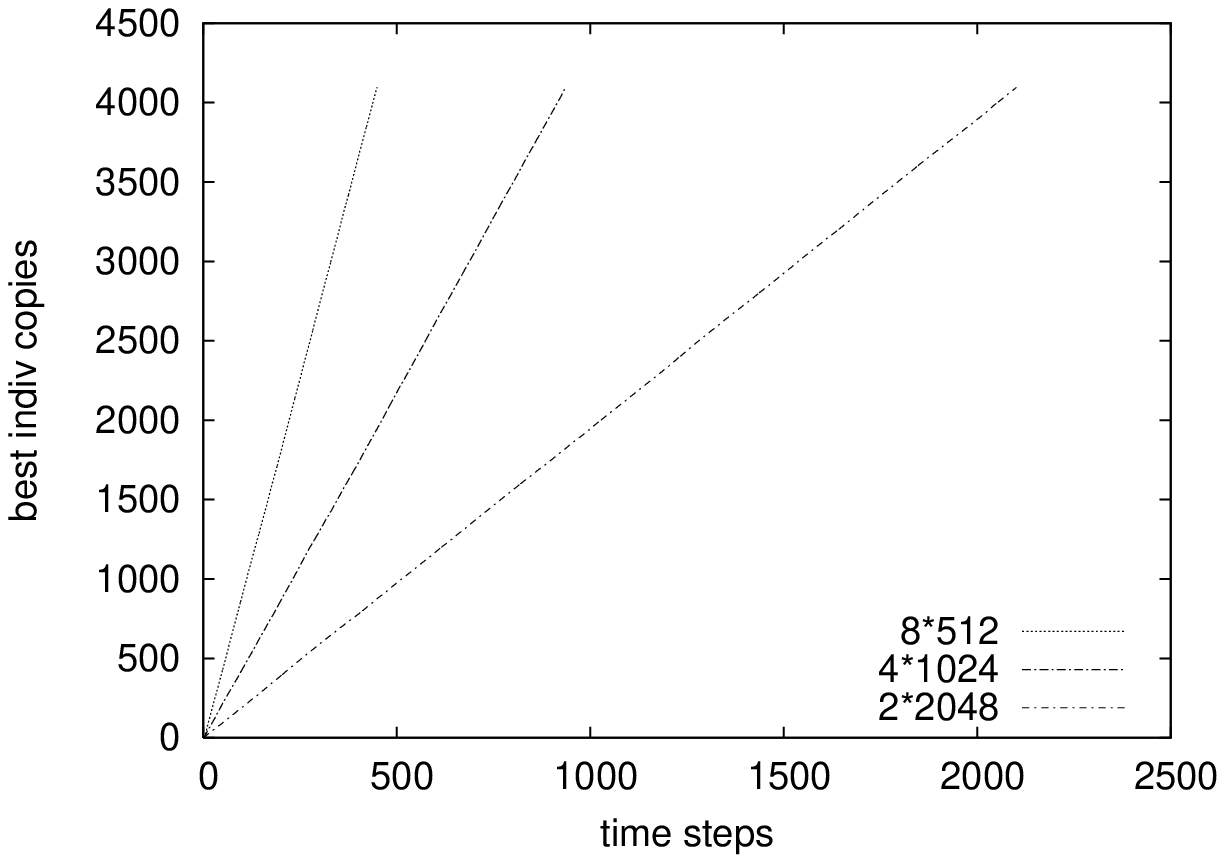} \\
\end{tabular}
\end{center}
\caption{Growth curves of the number of best individual copies $N(t)$ on different grid shapes.}
\label{fig-crois-tournoi}
\end{figure}

\begin{figure}[ht!]
\begin{center}
\includegraphics[width=6cm,height=6cm]{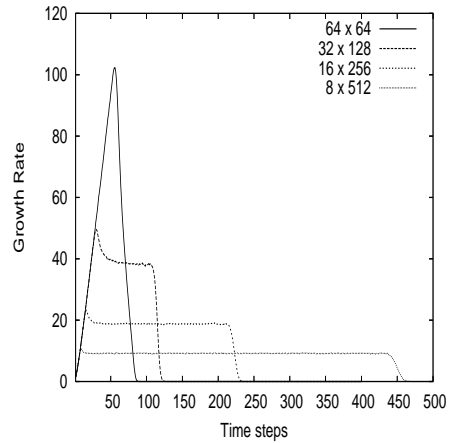} 
\end{center}
\caption{Growth rate against time steps for four rectangular grid shapes.}
\label{acrois_tournoi}
\end{figure}

\begin{figure}[h!]
\begin{center}
\begin{tabular}{ccc}
\includegraphics[width=1.5cm,height=6cm]{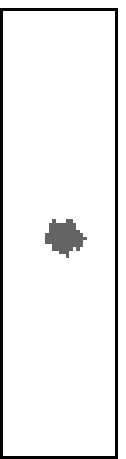} &
\includegraphics[width=1.5cm,height=6cm]{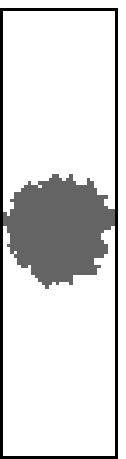} &
\includegraphics[width=1.5cm,height=6cm]{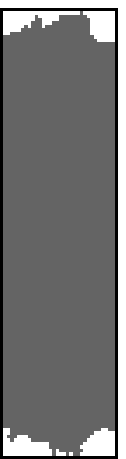}\\
(a) & (b) & (c) \\
\end{tabular}
\end{center}
\caption{Spreading of the best individual over a 32x128 grid}
\label{display_rec}
\end{figure}

\label{sec-EffectGridTopo}
We measure the relation between grid topology and selective pressure for 
rectangular grids where the population size is fixed to $4096$. We use the following grid
 shapes: $64\times64$; $32\times128$; $16\times256$; $8\times512$; $4\times1024$ and
 $2\times2048$ individuals. The selection strategy is a binary tournament. 
For each cell we randomly choose two individuals in its neighborhood. 
 The best one then replaces the individual of the cell on the grid if it is fitter or
 with probability 0.5 if fitnesses are equal. 
Figure \ref{fig-crois-tournoi} shows the average of $10^3$ independant runs of growth
 of $N(t)$ against generations; 
the takeover time is reached when $N(t)$ is equal to the size of the grid (see Table I). 
The average growth rate $\Delta(t)$, that is the number of new best individual copies
 per time unit, of these curves for four rectangular grid shapes
 ($64\times64$, $32\times128$, $16\times256$,$8\times512$) is plotted in 
Figure \ref{acrois_tournoi}. 
This figure helps us to understand the growth of $N(t)$. 
The growth rate $\Delta(t)$ is the same for all grids for the first generations. 
Then, the spreading speeds down to reach a constant speed for rectangular grids. 
This constant is $2 l p$ where $p$ is the probability of selecting the best
 individual when there is one copy of it in the neighborhood and $l$ the shortest
 side of the grid \cite{Giacobini2004}.
More accurately, $\Delta(t)$ decreases when $l$ is filled by copies of the best 
individual (see Figure \ref{display_rec}(b)).
Then, the speed becomes constant until the best individual has spread 
over to the other side (see Figure \ref{display_rec}(c)). 
 This explains why the $64\times64$ grid curve has no constant period: 
the two sides are filled at the same time. 
Finally, the growth rate falls down to zero as the best individual finishes conquering the grid. 
The results of the experiments we conducted are in agreement with
 E.Alba and J.Troya observations that narrow grid shapes induce low 
selective pressure \cite{AlbaT00}.
We will see in the next sections that this behavior can be observed with the AS too.

\clearpage

\begin{figure}[h!]
\begin{center}
\begin{tabular}{ccc}
\includegraphics[width=3cm,height=3cm]{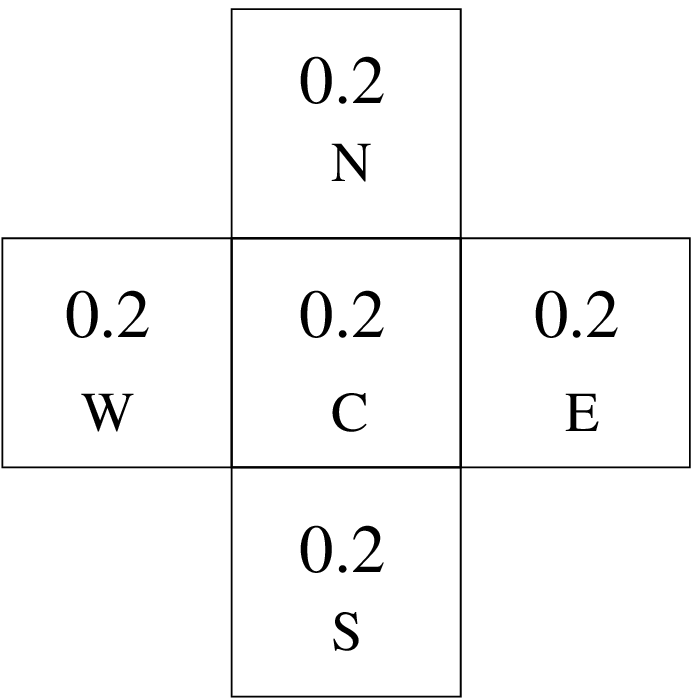} &
 &
\includegraphics[width=3cm,height=3cm]{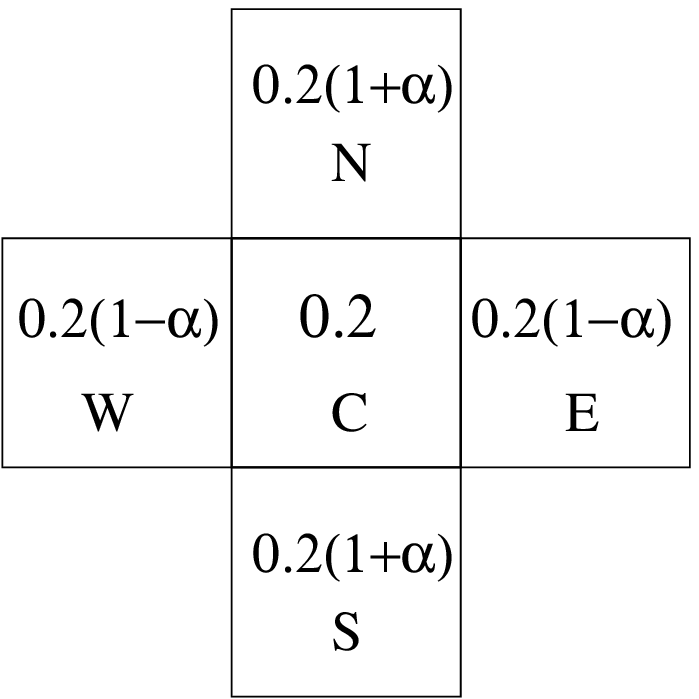} \\
\end{tabular}
\end{center}
\caption{Von Neumann and Von Neumann fuzzy Neighborhoods with probabilities to choose each neighbor}
\label{neighborhoods}
\end{figure}

\begin{table}[h!]
\begin{center}
\caption{Takeover time for different rectangular grid shapes.}
\begin{tabular}{|r|r r r|}
\hline
         & \multicolumn{3}{|c|}{Takeover Time} \\
Grid shape & Avg & Min & Max \\
\hline
$64\times 64$ & $83.4_{1.9}$ & $79$ & $87$ \\ 
$32\times 128$ & $117.8_{2.4}$ & $114$ & $123$ \\ 
$16\times 256$ & $225.0_{3.8}$ & $219$ & $232$ \\ 
$8\times 512$ & $449.7_{6.3}$ & $437$ & $463$ \\ 
$4\times 1024$ & $937.1_{9.9}$ & $921$ & $960$ \\ 
$2\times 2048$ & $2101.2_{29.9}$ & $2045$ & $2155$ \\ 
\hline
\end{tabular} 
\end{center}
\label{tab-takeover-rec}
\end{table}

\section{Anisotropic selection}
\label{section2}
This section introduces our contribution: the \textit{Anisotropic Selection} method 
where the neighbors of a cell may be selected with different probabilities.

\subsection{Von Neumann Fuzzy Neighborhood}
The Von Neumann neighborhood of a cell $C$ is defined as the ball of radius 1 in Manhattan 
distance centered at $C$. Using the \textit{Von Neumann Fuzzy Neighborhood}, we assign different
 probabilities to choose one cell in the neighborhood according to the directions (see Figure 
\ref{neighborhoods}). The probability $p_c$ to choose the center cell $C$ is
 set at $\frac{1}{5}$ as for 
Von Neumann neighborhood. Let $p_{ns}$ denote the probability to choose the cell $N$ or $S$ and
 $p_{ew}$ denote the probability to choose the cell $E$ or $W$.
Let $\alpha \in [-1;1]$ be the control parameter, the anisotropy degree. 
When $\alpha=-1$, we have $p_{ew}=1-p_c$ and $p_{ns}=0$, when 
$\alpha=0$ we have $p_{ns}=p_{ew}$ and when $\alpha=1$ we have $p_{ns}=1-p_c$ and $p_{ew}=0$. 
Thus, the probabilities 
$p_{ns}$ and $p_{ew}$ can be described as: \\
$$p_{ns} = \frac{(1-p_c)}{2}(1+\alpha)$$ \\
$$p_{ew} = \frac{(1-p_c)}{2}(1-\alpha)$$ \\
The case $\alpha=0$ correspond to the standard Von Neumann neighborhood
 ($p_{ns}=p_{ew}=\frac{2}{5}$)
 and $\alpha=1$ is the limiting case for fuzzy neighborhood where $p_{ns}=\frac{4}{5}$ and 
 $p_{ew}=0$. In the latter case, there is a vertical neighborhood with three neighbors
 only\footnote{
using the grid symmetry we will consider $\alpha \in [0;1]$ only}. 

\subsection{Definition}

The AS operator exploits the Von Neumann Fuzzy Neighborhood. It works as follows: 
for each cell $k$ individuals are selected accordingly to the
 probabilities $p_{ns}$, $p_{ew}$ and $p_{c}$ ($k$ stands in the range $[1,5]$) within the cell neighborhood. 
The $k$ individuals participate to a tournament and the winner replaces the old individual
 if it is fitter or with probability 0.5 if fitnesses are equal. 
The control parameter $\alpha$ is a measure of anisotropy: $\alpha=0$ corresponds to standard
 selection, and $\alpha=1$ is the limiting case with the utmost anisotropy. We conjecture that
 selective pressure decreases when anisotropy increases.

\section{Selective pressure and \\ Anisotropic Selection}
\label{section3}

In this section we study the relationship between selective pressure and AS.
First, we measure the takeover time for different anisotropy degrees, 
then we take the study further by considering the growth curve of the best individual.

\begin{figure}[ht!]
\begin{center}
\includegraphics[width=6cm,height=6cm]{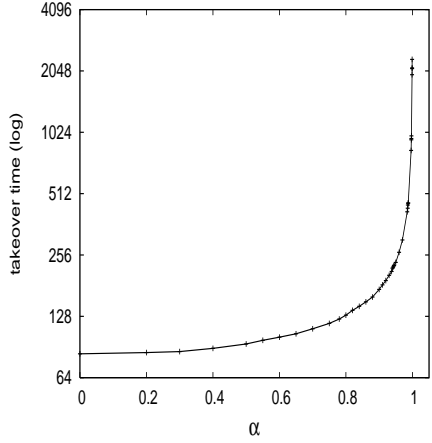}
\end{center}
\caption{Average of the takeover time as a function of the anisotropic degree $\alpha$.}
\label{fig-tak-fuzzy}
\end{figure}

We measure the effect of different anisotropy degrees on the takeover time. 
In our experiments, the anisotropic selection is based on tournament selection of
 size $k=2$ on the square grid of side $64$.
All the $4096$ cells are updated synchronously.
For each cell, we substitute the selected individual to the one already present in that cell, either systematically
 if the selected individual is fitter, or with probability 0.5 if fitnesses are equal.
For each value of $\alpha$, we perform $10^3$ independent runs.
When $\alpha=0$, no direction is privileged and AS is equivalent to the standard selection
 method.
When $\alpha=1$, only one direction is exploited, the grid can not be filled and the takeover
 time is not defined.
Figure \ref{fig-tak-fuzzy} shows the influence of AS on the takeover time :  
it increases with the parameter $\alpha$. 
These results are fairly consistent with our expectation that selection intensity
 decreases when the anisotropic degree increases. However, the correlation between takeover 
and anisotropy is not linear; it rapidly increases after the value $\alpha=0.9$.

\begin{figure}[h!]
\begin{center}
\begin{tabular}{c}
\includegraphics[width=6cm,height=6cm]{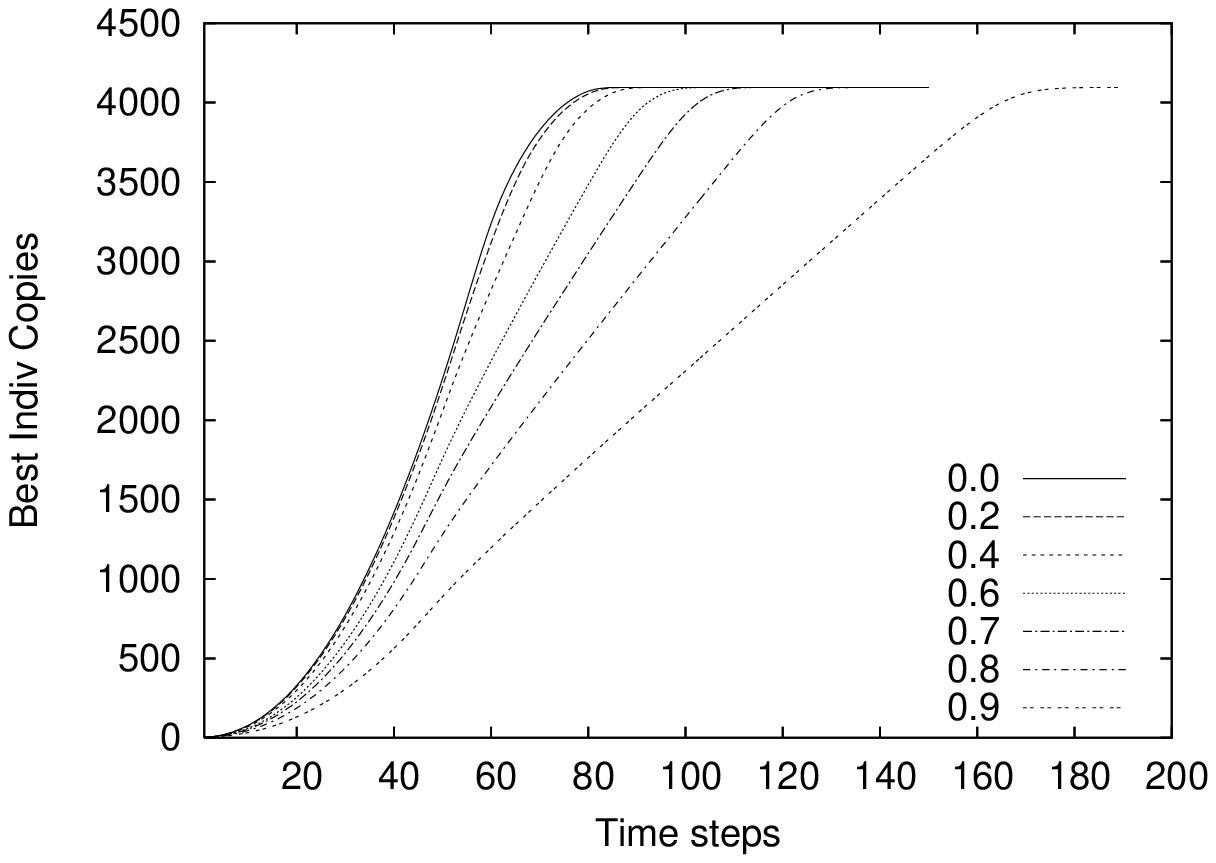} \\
(a) \\
\includegraphics[width=6cm,height=6cm]{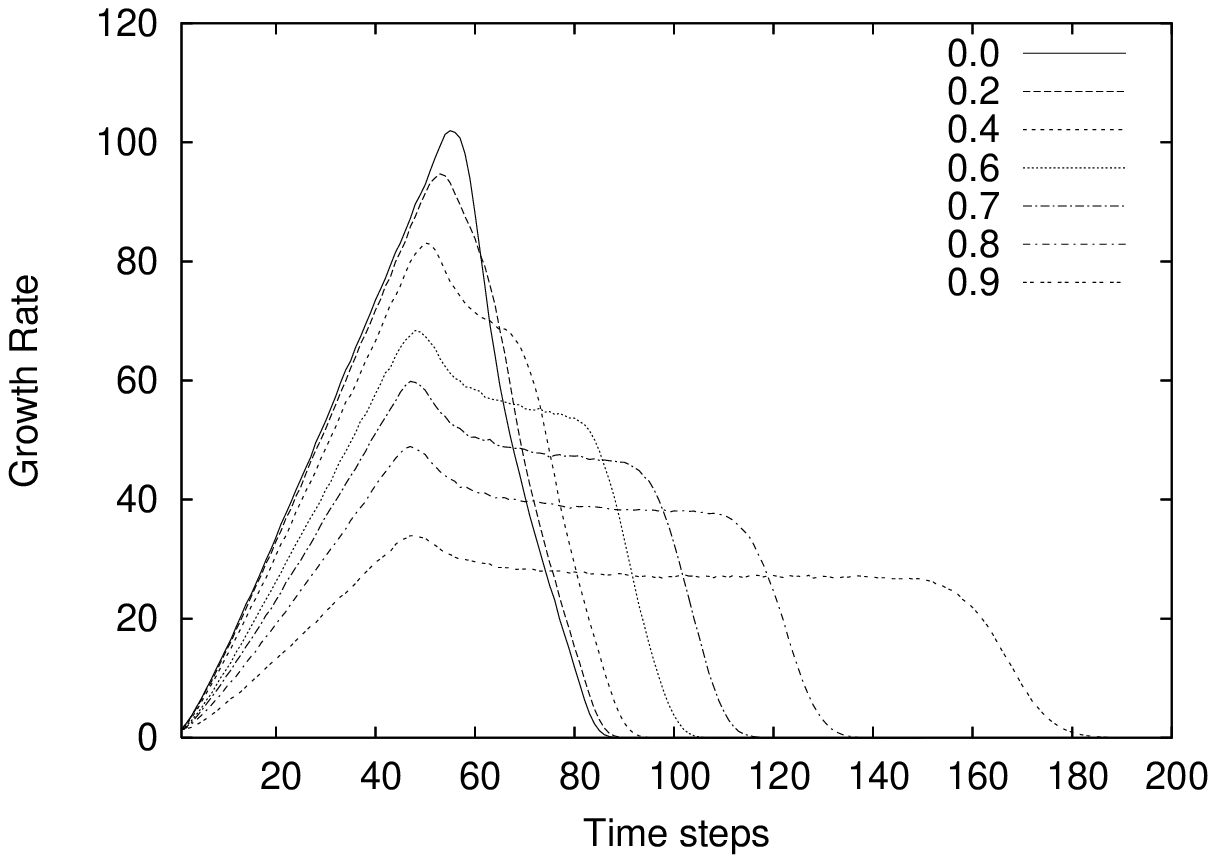} \\
(b) \\
\end{tabular}
\end{center}

\caption{Growth curves of $N(t)$ (a) and $\Delta(t)$ (b)  on a square grid for
 different anisotropic degrees $\alpha$.}
\label{fig-crois-fuzzy}
\end{figure}

\commentaire{
\begin{table}
\begin{center}
\begin{tabular}{|r|r r r|}
\hline
         & \multicolumn{3}{|c|}{Takeover Time} \\
$\alpha$ & Avg & Min & Max \\
\hline
$0.0$ & $83.8_{1.6}$ & $81$ & $87$ \\ 
$0.1$ & $84.3_{1.6}$ & $82$ & $88$ \\ 
$0.2$ & $85.0_{1.6}$ & $81$ & $88$ \\ 
$0.3$ & $86.0_{1.5}$ & $84$ & $90$ \\ 
$0.4$ & $89.2_{1.7}$ & $87$ & $93$ \\ 
$0.5$ & $93.7_{2.6}$ & $88$ & $98$ \\ 
$0.6$ & $101.2_{3.2}$ & $94$ & $106$ \\ 
$0.7$ & $111.2_{3.4}$ & $106$ & $119$ \\ 
$0.8$ & $129.7_{4.8}$ & $119$ & $138$ \\ 
$0.9$ & $173.4_{5.7}$ & $164$ & $185$ \\ 
\hline
\end{tabular} \\
\end{center}
\caption{Takeover time for anisotropic selection for different anisotropic degrees $\alpha$.}
\label{tab-takeover}
\end{table}
}

\eject
Figure \ref{fig-crois-fuzzy} shows the average curves of the growth of best individual
 (a) and its growth rate (b) as a function of time steps.
The shape of the curve is decomposed into three stages: 
in the first stage, the growth rate is almost proportional to the time steps and
the growth curve is approximatively a parabola.
In the second stage, this rate becomes roughly constant after a period of decrease and
the growth curve is almost linear.
In the last stage, the rate decreases linearly down to zero with a different slope than in the first stage.
The higher the anisotropic degree $\alpha$, the weaker the initial slope of the growth rate.
In the same way in the second stage, the slope of growth curve is smaller when
 $\alpha$ is higher.
Therefore, the selective pressure is lower when anisotropy is higher.

\begin{figure}[h!]
\begin{center}
\begin{tabular}{c}
\includegraphics[width=3.8cm,height=3.8cm]{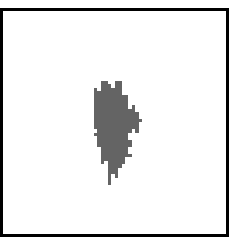} \\
(a) \\
\includegraphics[width=3.8cm,height=3.8cm]{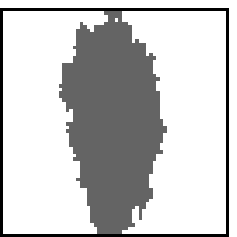} \\
(b) \\
\includegraphics[width=3.8cm,height=3.8cm]{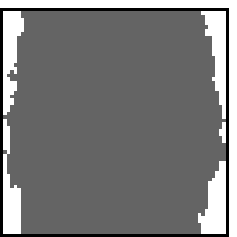} \\ 
(c) \\
\end{tabular}
\end{center}
\caption{Spreading of the best individual with $\alpha$=0.75}
\label{display}
\end{figure}

The three stages of the growth curve correspond to three periods in the spreading of the best
 individual on the square grid (see Figure \ref{display}).
During the first stage, the best individual spreads more in the privileged direction.
This period finishes, as described in Section \ref{sec-EffectGridTopo}, when
 a side of the grid is filled by best individual copies in the privileged 
direction (see Figure \ref{display}(b)).
During the second period, the best individual fills the second direction of the grid
until it has spread over a side of it in the less privileged direction
 (see Figure \ref{display}(c)).
The best individual front is sharp at the beginning, and becomes approximatively a 
horizontal line later.
The third time finishes to fill the grid.
Taking into account these three phases, one may be able to give the equation of
 the growth curve as in \cite{Giacobini2004}.

\begin{table}
\begin{center}
\caption{Values of $\alpha$ and $\frac{l}{L}$ for the same takeover time.
 Linear regression shows the relation between $\alpha$ and $\frac{l}{L}$ by the
 equation $\alpha$=$-0.999$$\frac{l}{L}$+$0.998$ with high correlation coefficient $-0.9999$}
\begin{tabular}{|l|l|l|}
\hline
\multicolumn{1}{|c|}{$l/L$} & 
\multicolumn{1}{|c|}{$\alpha$} &
\multicolumn{1}{|c|}{takeover time} \\
\hline
$0.000977$ & $0.99911$ & $2101$ \\ 
$0.003906$ & $0.99674$ & $939$ \\ 
$0.015625$ & $0.9864$ & $450$ \\ 
$0.0625$ & $0.944$ & $225$ \\ 
$0.25$ & $0.75$ & $118$ \\ 
$1.0$ & $0.0$ & $83$ \\
\hline 
\end{tabular} 
\end{center}
\label{mako}
\end{table}

\section{Anisotropic Selection vs. \\Rectangular Grid}
\label{section4}
Changing rectangular grid shape and tuning the anisotro\-pic degree are two methods for
 varying the selective pressure.
This section compares the two methods and shows in which way they are equivalent.

From the experimental results presented in Sections \ref{sec-takRG} and \ref{section3}, 
we compute the $\alpha$ parameter value for which we obtained the same takeover time as 
for one particular rectangular grid shape. 
In Table II we give the $\alpha$ and$\frac{l}{L}$ values and exhibit the correlation 
between these parameters.
We can see that $\alpha$ is proportional to the $\frac{l}{L}$ ratio.
So, according to the takeover time using $\alpha = 1 - \frac{l}{L}$, it is possible to
 have the same selective pressure using the two methods. 

\begin{figure}[ht!]
\begin{center}
\begin{tabular}{c}
\includegraphics[width=5cm,height=5cm]{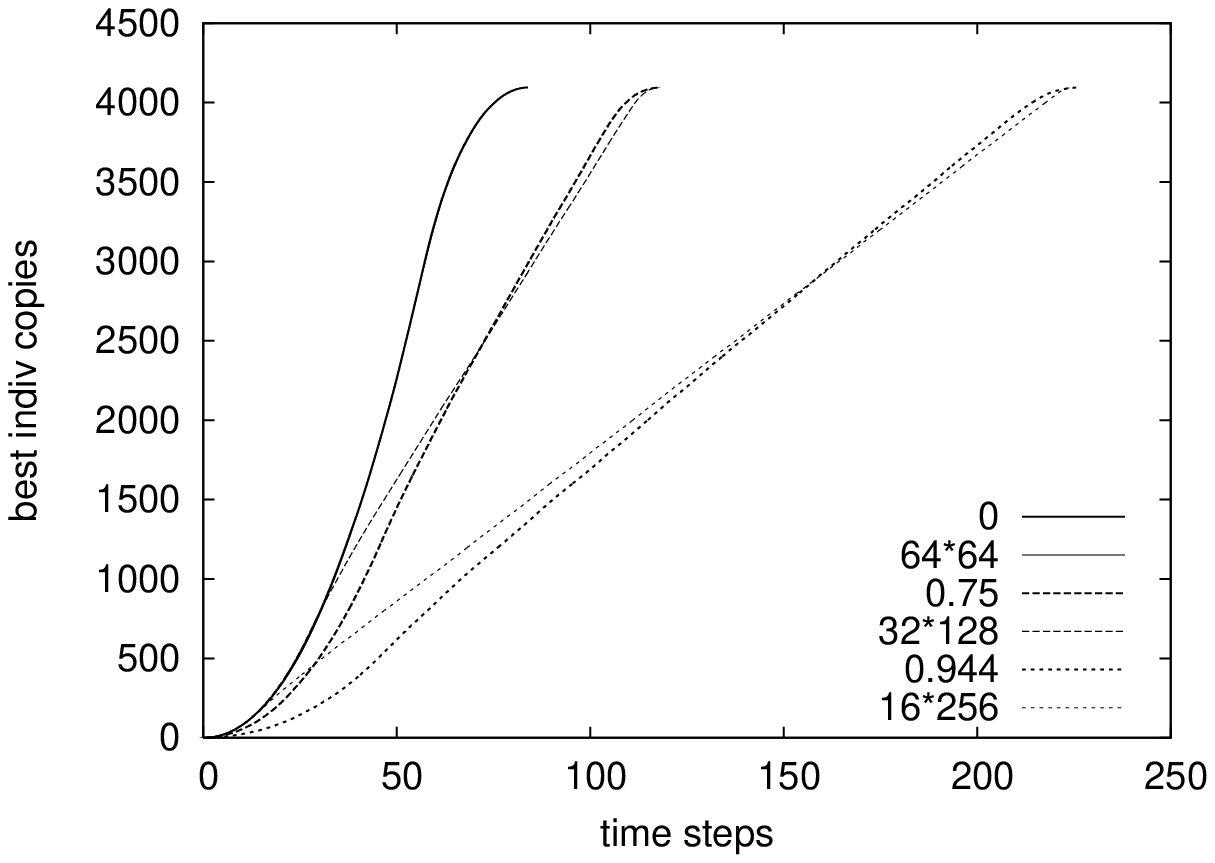} \\
\includegraphics[width=5cm,height=5cm]{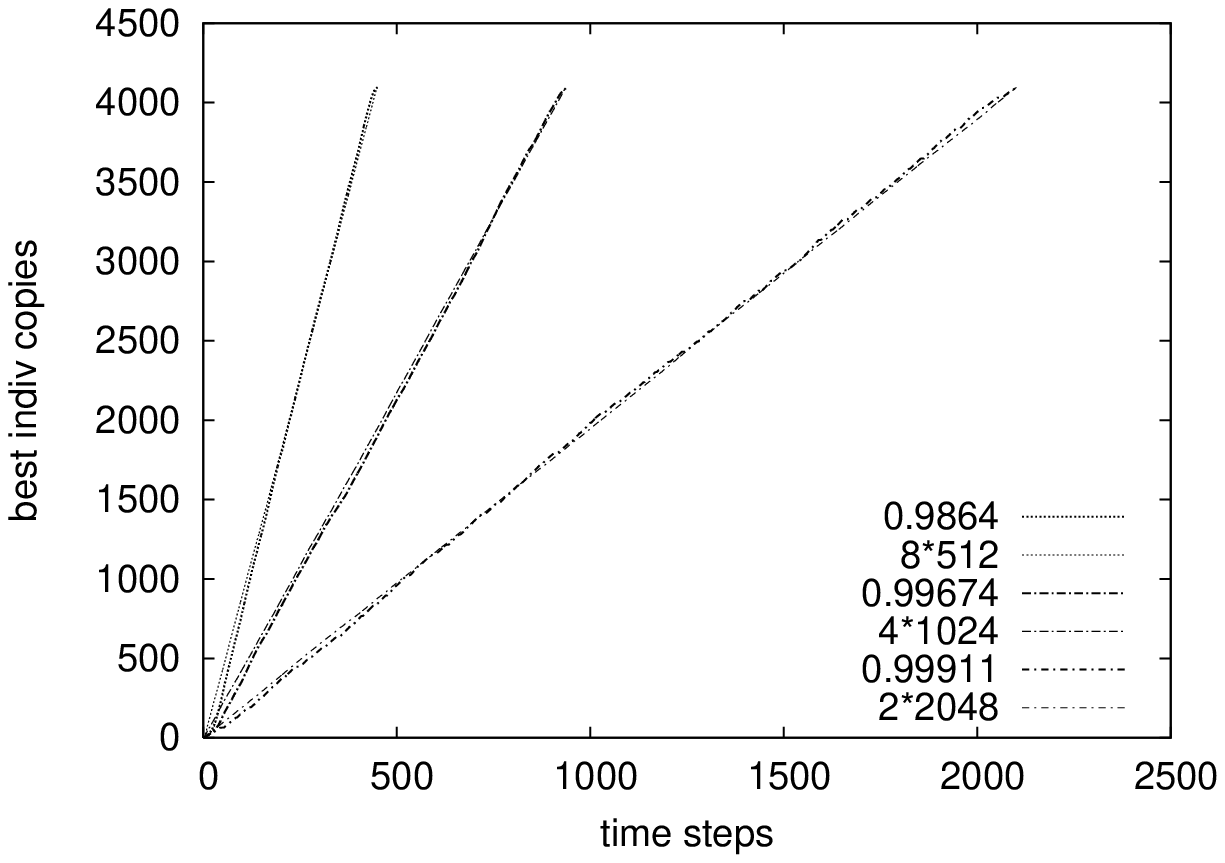} \\
\end{tabular}
\end{center}
\caption{Comparison between growth curves on different rectangular grids and on a 
square grid with the equivalent anisotropy degree.}
\label{merge}
\end{figure}

Figure \ref{merge} shows the mean growth curves of the best individual spreading
 against time steps
for all grid shapes and for the corresponding square grids using AS.
Although, using the relation $\alpha=1-\frac{l}{L}$, we found the same takeover time 
with rectangular grid shapes and with AS  
the selective pressure is applied in a different way for the two methods.
It is weaker during the first generations in the anisotropic case 
then it becomes slightly stronger, to finally fill the grid at the same takeover time.

\section{Anisotropic selection and \\niching}
\label{section5}
Many real optimization problems require the coexistence of diverse solutions 
during the search. In this section we show how the anisotropic selection is able to promote
niching. 

\subsection{Niching methods}

Niching methods have been proposed in the field of genetic algorithms to preserve 
population diversity
and to allow the GA to investigate many peaks in parallel. As a side
effect, niching prevents the GA from being trapped in local optima. Niching
methods are inspired from nature where species specialize themselves to different
 ecological niches in order to decrease the selective pressure they undergo.
Niching GA's tend to achieve a natural emergence of niches in the search
space. A niche is commonly referred to as an optimum of the domain, the
fitness representing the resources of that niche \cite{Sareni98}. Niching
methods are used to solve multimodal problems, and also in dynamic optimization \cite{Cedeno97}. 
For such problems a GA must maintain a diverse
population that can adapt to the changing landscape and locate better
solutions dynamically. There are different niching GA for panmictic population: 
sharing, crowding, etc. These methods are based on the concept of distance: 
sharing \cite{Golberg87} decreases the fitness according to the
number of similar individuals in the population, and with crowding,
replacement is performed considering the distance between solutions.

\subsection{Experimental results}
\begin{figure}
\begin{center}
\begin{tabular}{cccc}
Time steps &
$\alpha$=0 &
$\alpha$=0.75 &
$\alpha$=0.99674 \\ 
20 &
\fbox{\includegraphics[width=3cm,height=3cm]{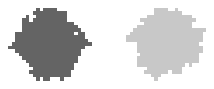}} &
\fbox{\includegraphics[width=3cm,height=3cm]{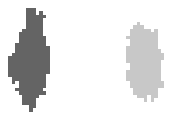}} &
\fbox{\includegraphics[width=3cm,height=3cm]{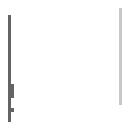}} \\
50 &
\fbox{\includegraphics[width=3cm,height=3cm]{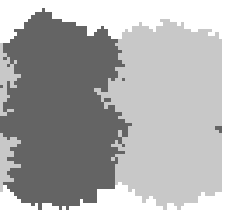}} &
\fbox{\includegraphics[width=3cm,height=3cm]{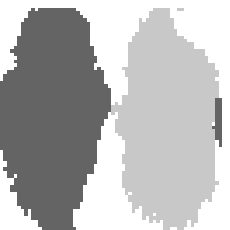}} &
\fbox{\includegraphics[width=3cm,height=3cm]{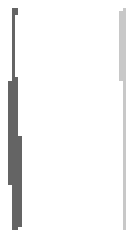}} \\
120 &
\fbox{\includegraphics[width=3cm,height=3cm]{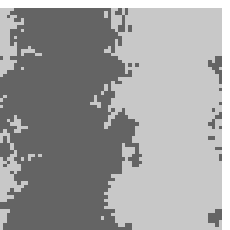}} &
\fbox{\includegraphics[width=3cm,height=3cm]{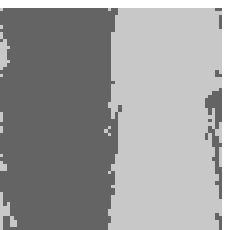}} &
\fbox{\includegraphics[width=3cm,height=3cm]{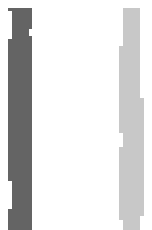}} \\
400 &
\fbox{\includegraphics[width=3cm,height=3cm]{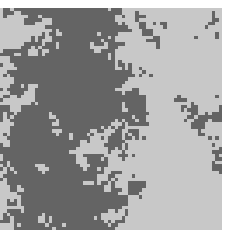}} &
\fbox{\includegraphics[width=3cm,height=3cm]{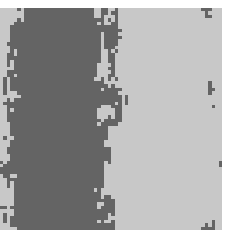}} &
\fbox{\includegraphics[width=3cm,height=3cm]{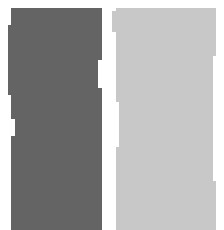}} \\
1000 &
\fbox{\includegraphics[width=3cm,height=3cm]{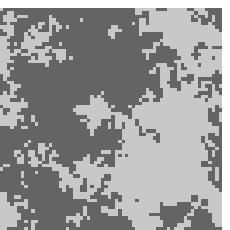}} &
\fbox{\includegraphics[width=3cm,height=3cm]{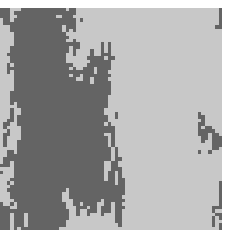}} &
\fbox{\includegraphics[width=3cm,height=3cm]{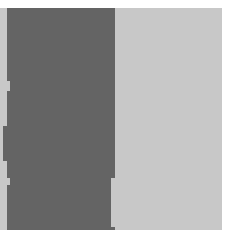}}\\
\end{tabular}
\caption{Spreading of two copies of the best individual}
\end{center}
\label{display_two}
\end{figure}

To show up to what extent anisotropic selection promotes niching, we have
conducted experiments where two solutions with the best fitness (here 1) are
placed on a $64\times64$ square grid at the initial generation. These solutions are farther
from each other in the least favored direction (here oriented horizontally).
 Figure \ref{display_two}
shows some snapshots of the spreading of the two bests over generations for
different anisotropic degrees. Cells in light grey (resp. dark grey) are
copies of the first best (resp. the second best), and all white cells  have
a null fitness value. Generations grow from top to bottom, and the
anisotropic parameter $\alpha$ increases from left to right. The left-hand
row ($\alpha=0$) represents standard binary tournament schema ; we observe
that standard selection is not able to maintain niches, after 1000
generations the grid is a mixture of the two optima. On the other hand, as
$\alpha$ increases, two stable frontiers between niches emerge. 
Hence AS increases \textit{cohesion} in each cells lineages.

\onecolumn

\twocolumn

\section{Test problem}
\label{section6}
We experiment a cGA using anisotropic selection on a Quadratic Assignment Problem (QAP): Nug30.
 Our aim here is not to obtain better results with respect to other optimization 
methods, but rather to observe the behavior of
a cGA with AS. In particular, we seek an optimal value for the anisotropy  degree.
\subsection{The Quadratic Assigment Problem}
The QAP is an important problem in both theory and practice. 
It was introduced by Koopmans and Beckmann
in 1957 and is a modal for many practical problems \cite{Koopmans57}.\\ 
The QAP can be described as the problem of assigning a set of facilities to
a set of locations with given distances between the locations and given flows between the 
facilities. The goal is to place the facilities on locations in such a way that the sum
 of the products between flows and distances is minimal.\\
Given $n$ facilities and $n$ locations, two $n \times n$ matrices $D=[d_{kl}]$ and $F=[f_{ij}]$
 where $d_{kl}$ is the distance between locations $k$ and $l$ and $f_{ij}$ the flow between 
 facilities $i$ and $j$, the objective function is: \\
\begin{displaymath}
\Phi = \sum_{i}\sum_{j}d_{p(i)p(j)}f_{ij}
\end{displaymath}
where $p(i)$ gives the location of facility $i$ in the current permutation $p$.\\
 Nugent, Vollman and Ruml suggested a set of problem instances of different sizes
 noted for their difficulty \cite{Nugent68}. The instances they suggested are known 
to have multiple local optima, so they are difficult for a genetic algortihm. 
We experiment our algorithm on their 30 variables instance called Nug30.

\subsection{Experiments}

We consider a population of 400 individuals placed on a square grid. Each individual represents
 a permutation of $\lbrace 1,2,...,30 \rbrace$. We need a special crossover 
that preserves the permutations:
\begin{itemize}
\item
  Select two individuals $p_1$ and $p_2$ as genitors.
\item
  Choose a random position $i$.
\item
  Find $j$ and $k$ so that $p_1(i) = p_2(j)$ and $p_2(i) = p_1(k)$.
\item
  swap positions $i$ and $j$ from $p_1$ and positions $i$ and $k$ from $p_2$.
\item
  repeat $n/3$ times this procedure where $n$ is the length of an individual.
\end{itemize}

This crossover is an extended version of the UPMX crossover proposed in \cite{Migkikh}.
The mutation operator consists in randomly selecting two positions from the individual
 and exchanging these positions. The crossover rate is 1 and we perform one mutation per
 individual in average.\\
We consider 500 runs for each anisotropy degree. Each run stops after 1500 
generations. 

\begin{figure}[ht!]
\begin{center}
\includegraphics[width=6cm,height=6cm]{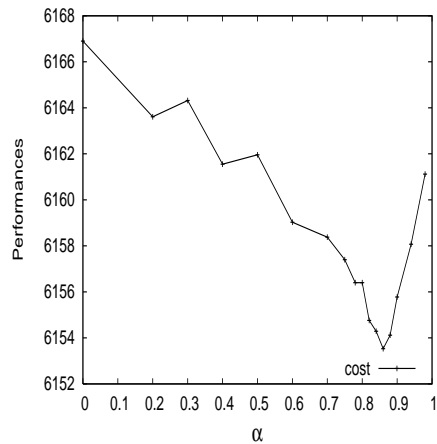}
\end{center}
\caption{Average costs as a function of $\alpha$.}
\label{alpha-perf}
\end{figure}

Figure \ref{alpha-perf} shows the average performance of the algorithm towards $\alpha$: 
for each value of $\alpha$ we average the best solution of each run.
 Performances are growing with $\alpha$ and then fall down as $\alpha$ is getting 
closer to its limit value.
This curve shows the influence of the selective pressure on the performances and how important 
 it is to control it accurately.  

The best average performance is observed for $\alpha=0.86$, which corresponds to a good 
exploration/exploitation tradeoff. In the neighborhood of this optimal value the
 algorithm favors propagation of good solutions in the vertical direction with 
few interactions on the left or right sides. This kind of dynamics is well
 adapted to multi-modal problems as we can reach local optima on each 
columns of the grid and then migrate them horizontally to find new solutions. 

Performances would probably improve if the selective pressure did not remain 
static during the search process. As in \cite{Alba05}, we can define some
 criteria to self-adjust the anisotropy degree along generations. 
Furthermore, we can assign a different anisotropy degree to each cell of the 
 grid, so that we can determinate criteria to self-adjust selective pressure locally upon measures on 
neighborhoods. 

\section*{Conclusions and Perspectives}

This paper presents a new selection scheme in
cellular genetic algorithms. The main objective is to control  the exploration/exploitation tradeoff in a flexible
way. We propose to exploit the
cellular GA characteristics to promote diversity during a genetic search
process. Previous studies on cGAs selected structural parameters, as
neighborhood or grid shape, to tune the selective pressure. The main
drawback of these techniques is that altering a structural parameter entails
a
deep change in the way we deal with the problem. The new selection scheme we
suggest is based on \textit{fuzzy neighborhood} where a cell
is chosen according to different probabilities. In order to favor one
direction
rather than the other one, \textit{anisotropic selection} chooses
individuals
in fuzzy neighborhood. Experiments performed in order to establish relation between
the takeover time and the degree of anisotropy are consistent with our
expectation that selection pressure decreases with the degree of anisotropy.
Analysis of the growth curves allows to distinguish three different phases
in the diffusion process. Experimental results establish linear
correlation in takeover between AS and cGA using rectangular grid. Then
we point out capabilities of AS to promote the emergence of niches.
Finally, using a cGA with AS on a QAP we have shown the existence of
 an anisotropic optimal value of $\alpha$ such that the best average performance
 is observed.

This paper is a preliminary investigation and a more extensive analysis must
be made to confirm that equilibrium between exploration and exploitation
makes AS a good technique for complex problems in static or dynamic
environments. Future work should address the following issues: comparison between AS
and changes in the neighborhood shape and size, measuring AS effects with cGA
using mutation and cross\-over, change the balance of directions dynamically.
The latter point is an important feature: by tuning the control parameter
$\alpha$, it would be possible to make the algorithm to self-adjust the
selective pressure, depending on global or local measures. This
adaptive ability has two important advantages: first, parameter $\alpha$
may vary in a continuous way, second, variations of this parameter have affect neither on the grid topology
 nor on the neighborhood shape.
Such self-adaptive algorithms have been studied in previous works, but they
need to change the grid topology to control the selective pressure \cite{Alba05}
, which means 
it is uniform in the grid. AS allows different propagation speeds on each area
of the grid, promoting diversification and intensification (exploitation) at the 
same time on different spots.  
 In general, we have to continue investigation of \textit{Anisotropic Selection}
to assess its validity and generality.

\bibliographystyle{abbrv}

\end{document}